\documentclass[sigconf, nonacm]{acmart}
\AtBeginDocument{%
  }

\setcopyright{acmlicensed}
\copyrightyear{2018}
\acmYear{2018}
\acmDOI{XXXXXXX.XXXXXXX}
\acmConference[Conference acronym 'XX]{Make sure to enter the correct
  conference title from your rights confirmation email}{June 03--05,
  2018}{Woodstock, NY}
\acmISBN{978-1-4503-XXXX-X/2018/06}

\usepackage{adjustbox}
\usepackage{booktabs,tabularx,threeparttable, placeins}
\begin{document}

\title{Assist-As-Needed: Adaptive Multimodal Robotic Assistance for Medication Management in Dementia Care}

\author{Kruthika Gangaraju}
\email{kgangaraju@wpi.edu}
\affiliation{%
  \institution{Worcester Polytechnic Institute}
  \city{Worcester}
  \state{Massachusetts}
  \country{USA}
}

\author{Tanmayi Inaparthy}
\email{tinaparthy@wpi.edu}
\affiliation{%
  \institution{Worcester Polytechnic Institute}
  \city{Worcester}
  \state{Massachusetts}
  \country{USA}
}

\author{Jiaqi Yang}
\email{jyang6@wpi.edu}
\affiliation{%
  \institution{Worcester Polytechnic Institute}
  \city{Worcester}
  \state{Massachusetts}
  \country{USA}
}

\author{Yihao Zheng}
\email{yzheng8@wpi.edu}
\affiliation{%
  \institution{Worcester Polytechnic Institute}
  \city{Worcester}
  \state{Massachusetts}
  \country{USA}
}

\author{Fengpei Yuan}
\email{fyuan3@wpi.edu}
\affiliation{%
  \institution{Worcester Polytechnic Institute}
  \city{Worcester}
  \state{Massachusetts}
  \country{USA}
}

\renewcommand{\shortauthors}{Gangaraju et al.}

\begin{abstract}
People living with dementia (PLWDs) face progressively declining abilities in medication management—from simple forgetfulness to complete task breakdown—yet most assistive technologies fail to adapt to these changing needs. This one-size-fits-all approach undermines autonomy, accelerates dependence, and increases caregiver burden. Occupational therapy principles emphasize matching assistance levels to individual capabilities: minimal reminders for those who merely forget, spatial guidance for those who misplace items, and comprehensive multimodal support for those requiring step-by-step instruction. However, existing robotic systems lack this adaptive, graduated response framework essential for maintaining PLWD independence. We present an adaptive multimodal robotic framework using the Pepper robot that dynamically adjusts assistance based on real-time assessment of user needs. Our system implements a hierarchical intervention model progressing from (1) simple verbal reminders, to (2) verbal + gestural cues, to (3) full multimodal guidance combining physical navigation to medication locations with step-by-step verbal and gestural instructions. Powered by LLM-driven interaction strategies and multimodal sensing (verbal, touch-screen, and camera), the system continuously evaluates task states to provide just-enough assistance—preserving autonomy while ensuring medication adherence. We conducted a preliminary study with healthy adults and dementia care stakeholders in a controlled lab setting, evaluating the system's usability, comprehensibility, and appropriateness of adaptive feedback mechanisms. This work contributes: (1) a theoretically-grounded adaptive assistance framework translating occupational therapy principles into HRI design, (2) a multimodal robotic implementation that preserves PLWD dignity through graduated support, and (3) empirical insights into stakeholder perceptions of adaptive robotic care. Our findings establish foundations for future trials with PLWDs, advancing toward truly person-centered robotic dementia care that augments rather than replaces human caregiving.
\end{abstract}



\received{20 February 2007}
\received[revised]{12 March 2009}
\received[accepted]{5 June 2009}

\maketitle

\section{Introduction}
\label{sec:intro}

The population of adults aged 65 and above has increased rapidly with rising life expectancy globally. In the United States alone, there were an estimated 54.1 million older adults in 2019, a number projected to exceed 80 million by 2040 \cite{administration20212020}. As adults 65 and older now comprise roughly one in seven Americans, age-related disorders have become more prevalent. Cognitive decline occurs due to aging \cite{salthouse2009does}, and when it gets severe enough to affect independence in activities of daily living (ADLs), it is classified as dementia \cite{mckhann2011diagnosis}. Approximately 10 million new dementia cases are reported worldwide each year, with Alzheimer’s disease (AD)—the most common cause—accounting for roughly 60–80\% of cases \cite{alzheimer2024alzheimer}. In the United States, an estimated 6.9 million adults aged 65+ were living with AD or AD-related dementias (ADRD) in 2024. AD/ADRD are progressive neurodegenerative conditions characterized by cognitive, behavioral, and/or motor impairments \cite{trojsi2018behavioral} that reduce independence and quality of life \cite{andersen2004ability}. People living with dementia experience declines in memory, language, and problem-solving that hinder activities of daily living (ADLs), often necessitating assistance to initiate and complete tasks \cite{andersen2004ability}. One of the commonly faced issue by PLWDs is progressively lose the ability to manage medications, from minor forgetfulness to complete task breakdown \cite{arlt2008adherence}, which in turn imposes significant strain on the caregiver. Although no cure exists to date, research increasingly targets early diagnostic approaches and supportive interventions. Among these, artificial intelligence (AI) and socially assistive robotics (SAR) have emerged to support cognitive, emotional, and physical aspects of care for individuals with AD/ADRD \cite{hirt2021social}, with the potential to reduce caregiver burden and to detect and mitigate ADL difficulties. \\
SAR is effective in providing the necessary reminders and guidance required for dementia caregivers. While it is effective, it also requires powerful, reliable, and trustworthy AI that can identify forgetful moments and deliver appropriate, timely assistance at the right time and place \cite{hazzan2022family, xie2020artificial}. The goal of having SAR for caregiving is to prompt the PLWD to initiate and complete their task by providing \emph{minimal yet effective} support without taking away their independence, which may contribute to slower decline of cognitive abilities and disease progression \cite{morris2022social}. While early robotic assistants such as  \cite{pineau2003towards, graf2004care} and more recent systems such as RAMCIP demonstrated that mobile robots can issue reminders and even physically deliver objects \cite{kostavelis2018ramcip}, these systems largely provide fixed, one-size-fits-all interventions. They do not systematically adjust assistance to a person’s fluctuating cognitive and functional abilities, nor do they operationalize the \textbf{assist-as-needed} principles long established in occupational therapy. This gap limits their capacity to preserve autonomy and dignity as dementia progresses.

Additionally, existing robotic systems for dementia care typically rely on bespoke speech recognition and rule-based interaction flows, restricting their ability to handle ambiguous or unexpected human behaviors ~\cite{wang2017robots}. Even when speech, vision, and navigation are present, modalities are often stitched together with brittle heuristics rather than a unified, uncertainty-aware decision layer, making cross-modal conflicts and sensor dropouts hard to manage. They also rarely involve care professionals or stakeholders early in the design process, leaving uncertainty about acceptability and appropriateness.
  \begin{figure}[ht!]
  \centering  \includegraphics[width=\columnwidth]{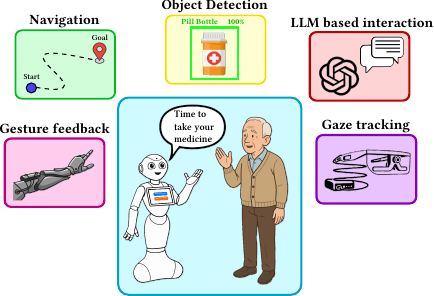}
\caption{Adaptive \textit{Assist-As-Needed} for people with dementia, involving multimodal perception and feedback}
 \label{fig:mindcare}
\end{figure}
We address these limitations by introducing one of the first robotic system to explicitly translate occupational therapy’s adaptive "assist-as-needed" framework as a concrete, testable human–robot interaction pipeline for medication management in dementia care. Running on the Pepper platform, our system combines GPT‑4o \cite{hurst2024gpt} for dialogue and task reasoning, Whisper \cite{radford2022robustspeechrecognitionlargescale} for speech recognition, and YOLOv11 \cite{yolo} for visual understanding of medication artifacts; it fuses these with gesture feedback, an on‑device touchscreen, and autonomous navigation (\autoref{fig:mindcare}). This approach preserves autonomy and dignity, while ensuring support when needed. We also evaluate the system with healthy adults and dementia care stakeholders to validate usability, comprehensibility, and appropriateness of adaptive feedback mechanisms. \\
Guided by the goal of "HRI Empowering Society", supporting independence, dignity, and equitable access to care, we investigate two research questions: \textbf{RQ1)} How effective is the involvement of a social robot in supporting medication adherence relative to a verbal-only baseline? \textbf{RQ2)} What level of assistance (e.g., minimal cues vs.\ step-by-step guidance) do users prefer, and how does it affect perceived workload and autonomy? To answer these questions, we compare our adaptive, multimodal system against a \emph{verbal‑only} baseline that offers reminders and reassurance but provides no stepwise instruction, on‑device visuals, pointing/gestures, or shared control. We operationalize adherence via task completion, complement these with efficiency measures (time on task; number of interactions), and assess experience using standardized ratings of workload and perceived autonomy. Our a priori expectation is that the adaptive, embodied approach will increase adherence and reduce workload \emph{without} undermining autonomy—and may improve it by offering the \emph{least assistance necessary} at each step.

In doing so, our work speaks to empowerment at three levels: (i) for people living with dementia, by preserving agency through graded, user‑controlled assistance; (ii) for caregivers and clinicians, by making assistance transparent, configurable, and reviewable; and (iii) for health systems, by demonstrating an interpretable decision policy that can generalize beyond scripted lab tasks. We report quantitative and qualitative outcomes from healthy adults and dementia‑care stakeholders to evaluate usability, comprehensibility, and the appropriateness of adaptive feedback, and to surface design principles for dignifying, autonomy‑preserving HRI in medication management.

\section{Related Work}
Prior work has explored assistive robots as companions for people living with dementia (PLWDs)~\cite{joranson2016change, liao2023use}, examined the acceptance of social robots in care settings~\cite{ke2020changes}, and investigated their use in rehabilitative therapy~\cite{cruz2020social}. Despite this progress, usability and effectiveness for concrete assistance with activities of daily living (ADLs) in real‑world settings remain under‑examined~\cite{ghafurian2021social}.

Within dementia care, robots have been deployed for task‑level prompting and graded assistance. Wang et al.~\cite{wang2017robots} implemented a teleoperated socially assistive robot that delivered need‑based assistance and reported acceptance from both PLWD and caregiver perspectives; however, the system depended on a human teleoperator rather than automated decision making. Mobile platforms have also demonstrated physical assistance: the Nursebot “Pearl”~\cite{pineau2003towards} integrated an actuated head, onboard sensors, speech synthesis, and a touchscreen to provide navigational support and time‑contingent medication reminders for older adults, while Care‑O‑bot~\cite{graf2004care} explored manipulation for pick‑and‑place of common indoor objects to assist older adults and people with physical disabilities. Neither system centered social interaction or the principle of providing help only when needed to preserve PLWDs’ independence. More recently, Pepper has been used in dementia‑relevant task contexts: Yuan et al.~\cite{yuan2022assessment, yuan2024social} teleoperated Pepper to assist PLWDs with PASS tasks, examining assistance levels and underscoring the need for adaptive, personalized interactions.

Across this literature, a recurring theme is the integration of multimodal feedback to support smooth task execution while reducing cognitive load. Typical modalities include (i) auditory prompts (natural‑language reminders, stepwise instructions, confirmations), (ii) visual cues on an integrated screen (checklists, icons, timing indicators), and (iii) embodied signals such as pointing, head orientation and gaze, and purposeful proxemics (navigating to the relevant location or object), with manipulation when available. Effective systems blend these channels to enable graded cueing—from high‑level reminders to fine‑grained step guidance—clarify environmental referents, and support error recovery, while preserving user autonomy and accommodating individual preferences.
\begin{figure*}[t]
  \centering
  \includegraphics[width=\textwidth]{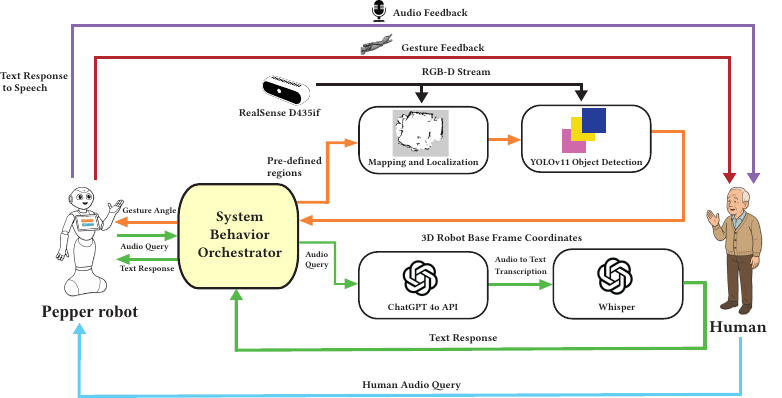}
\caption{Overview of the framework for medication management. The arrows indicate the flow of information.}
 \label{fig:overview}
\end{figure*}
\section{Methodology}
We developed an adaptive multimodal robotic framework that integrates multimodal perception (audio, touchscreen, and camera/visual, and eye-tracking) with an LLM-backed dialogue manager to deliver adaptive, multimodal assistive guidance during a medication-taking routine for people with dementia. Implemented on the Pepper humanoid robot, the system combines verbal and  gestural feedback - including gaze alignment and arm pointing - to support task completion while preserving autonomy. \autoref{fig:overview} illustrates the system framework. In our implementation, the robot localizes against a pre‑built map of the test environment and uses the standard ROS navigation stack to move to a designated region of interest (ROI). Regions of interest (ROIs) are predefined, manually labeled map locations where the pill bottle may be found. We selected this arbitrarily to simulate real-world scenario where the PLWDs could possibly leave their medication. Upon arrival to the ROI, the perception modules detect the pill bottle and water bottle, estimate the target’s 3‑D pose, and generate embodied cues while concurrently delivering verbal and on‑screen instructions. A system behavior orchestrator ensures the correct flow of information during the entire process. The navigation and object‑detection stack was designed with complex deployment settings in mind (e.g., home environments or assisted‑living centers). In such cases, multiple ROIs can be defined; the robot can visit each as needed, check for the pill bottle, and then guide the user to the exact location for the next steps. The system also incorporates an eye-tracking sensor to capture participants’ gaze patterns; although the full adaptive logic (e.g., prompting when prolonged gaze is detected) has not yet been activated, this capability was used to collect gaze data from participants to inform future development.

\vspace{-2mm}
\subsection{Enhanced Robotic Platform with Integrated Sensors}
\begin{figure}[!htp]
  \centering  \includegraphics[width=0.65 \columnwidth]{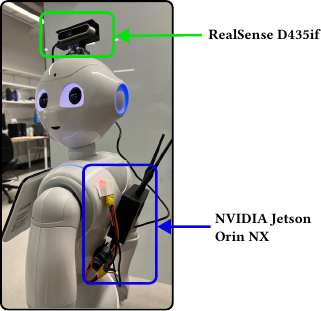}
\caption{Hardware components mounted on Pepper robot}
 \label{fig:hardware}
\end{figure}

In order to implement our framework, we made use of a SAR. Pepper robot is a widely used social robot in healthcare and hospitality sectors. The Pepper robot stands at 1.21-m tall and features 17 joints for graceful and expressive body movements. Pepper supports expressive non-verbal communication (gestures, posture, head gaze) and speech synthesis, and integrates RGB cameras, microphones, sonars/IR proximity sensors, and an on-chest tablet for visual prompts.  Although Pepper is suitable for immersive, engaging human-robot interaction, Pepper’s native perception stack and depth sensing capabilities are limited for robust navigation and object detection. Additionally, its onboard computational power is insufficient for time-critical assistance and interaction with people with dementia.
To address these limitations, we added an NVIDIA Jetson Orin NX module on its backplate for onboard inference and mounted an Intel RealSense D435if camera on top of its head, (as shown in \autoref{fig:hardware}). The RealSense camera provides RGB–D streams, and the Orin runs SLAM preprocessing and object detection at the edge. Pepper’s on-board computer communicates with the Orin over Wi‑Fi using ROS and a flask server.

\subsection{Navigational Guidance}
People living with dementia (PLWDs) may forget where medications or related items (e.g., water bottles) are placed, so the system autonomously visits likely storage locations to ground subsequent detection and guidance. We implement 2-D SLAM with \texttt{slam\_gmapping}, using the head-mounted RealSense depth stream converted to a virtual laser via \texttt{depthimage\_to\_laserscan}; an occupancy grid is built during an initial exploration pass \cite{grisetti2007gmapping}. For localization on the saved map, we use Adaptive Monte Carlo Localization (\texttt{amcl}), which maintains an online-adaptive particle filter over robot pose \cite{fox2001amcl}. Path execution is managed by \texttt{move\_base} \cite{quigley2009ros}: the global planner (Dijkstra/A* over the static costmap) provides routes, and the local planner implements the Dynamic Window Approach to select feasible velocity commands \cite{fox1997dwa}. Layered costmaps (\texttt{costmap\_2d}) combine a static layer (map), an obstacle layer (virtual-laser/depth obstacles), and an inflation layer to maintain clearance. Navigation goals are specified as regions of interest (ROIs)—predefined map coordinates where the pill bottle may be found. At run time, the system behavior orchestrator (see \autoref{fig:overview}, orange line) sends ROI waypoints sequentially to \texttt{move\_base}; Once the robot reaches an ROI, it pauses motion and triggers the YOLO detection module (see ~\autoref{sec:perception}). If the model does not detect the object, the robot moves to the next ROI. Velocity commands are published on \texttt{/cmd\_vel} to the Pepper base controller, positioning the robot for the subsequent joint-attention and dialogue phases.

\subsection{Multimodal Perception}
\label{sec:perception}
Multimodal perception enables robots to operate in complex, human-centered settings by fusing heterogeneous signals-vision, 
speech / language, and touch-into a coherent model for interaction \cite{mollaret2016multi, zhao2025multimodal}. In our system, camera-based perception is integrated with on-device touchscreen input and microphone audio to ground dialogue in the physical context.
\paragraph{Object Detection Module} After the robot localizes on a prebuilt map and navigates to the region of interest (ROI), it must reliably detect medication-related objects (pill bottle, water bottle) under variation in lighting, viewpoint, and clutter, while meeting real-time constraints on the Jetson. For these reasons, we use YOLOv11 \cite{yolo} for object detection initialized from COCO weights \cite{coco}. We collected 1{,}200 images spanning diverse viewpoints, lighting conditions, and levels of clutter, and applied standard geometric and photometric augmentations to improve robustness. At the ROI, the system behavior orchestrator pans the head from $-30^\circ$ to $+30^\circ$ and acquires RGB frames at regular intervals; detection is run on each frame, and upon a positive detection we extract the bounding box coordinates. To obtain a 3D target pose, we refine the detection to a per-pixel mask within the box using the aligned depth image: we compute the median depth $Z_m$ inside the box, retain pixels whose depth lies within a narrow band around $Z_m$ (depth-guided foreground extraction), and keep the largest connected component intersecting the box center as the segmented pill bottle. For each mask pixel $(u,v)$ with depth $Z(u,v)$ and camera intrinsics $(f_x,f_y,c_x,c_y)$, we back-project to the camera frame using the pinhole model $X=\tfrac{(u-c_x)Z}{f_x},\;Y=\tfrac{(v-c_y)Z}{f_y},\;Z=Z$, stack $p_{\mathrm{cam}}=[X,Y,Z,1]^\top$, and transform to the robot base via the calibrated extrinsics $p_{\mathrm{base}}=T_{\mathrm{base}\leftarrow\mathrm{cam}}\,p_{\mathrm{cam}}$. The resulting point cloud in the base frame is used to estimate a pointing direction: we take a small patch around the cloud centroid $\bar{p}$, fit a plane $ax+by+cz+d=0$ by least squares (normal $\mathbf{n}$ given by the eigenvector of the local covariance with the smallest eigenvalue), and orient the normal opposite the camera optical axis ($\mathbf{n}\cdot\hat{\mathbf{z}}_{\mathrm{cam}}<0$). Finally, we compute the pointing command from the vector $\mathbf{d}=\bar{p}_{\mathrm{base}}-\mathbf{o}_{\mathrm{arm}}$ to obtain yaw and pitch, $\psi=\operatorname{atan2}(d_y,d_x)$ and $\theta=\operatorname{atan2}(d_z,\sqrt{d_x^2+d_y^2})$, which drive a brief arm gesture that directs user attention to the bottle (see \autoref{fig:overview}, orange data flow).
\paragraph{Audio Perception Module}
The Pepper robot’s built-in microphones capture users’ spoken input. To initiate capture, the user presses the “Record” button presented on the robot’s tablet (push-to-talk). The resulting audio stream is passed to the Whisper model for automatic speech recognition (ASR), yielding a text transcript of the user’s response. This transcript is then forwarded to the downstream dialogue component (e.g., GPT‑4o in our system) to interpret intent and generate the appropriate response. The audio module therefore provides a simple interface—explicit user‑controlled recording, ASR transcription via Whisper, and text handoff to the interaction policy—so that higher‑level modules can adapt assistance without relying on always‑listening input (see \autoref{fig:overview}, blue and green data flow).
\subsection{Multimodal Feedback}

Motivated by prior HRI findings that complementary modalities improve comprehensibility and engagement \cite{nault2024socially, bolarinwa2019use} -particularly for older adults and PLWD - we designed multimodal robotic feedback to minimize cognitive load and disambiguate references in the environment. Our design takes advantage of Pepper's native features (speech synthesis, head / arm gestures, base orientation, and chest-mounted tablet) and delivers signals in two phases - verbal and deictic. In addition to these feedback modalities, the robot also provides \textit{navigational feedback} by leading the user to the location of the pill bottle.
\paragraph{Gestural Feedback} Once the pill bottle is detected, the robot establishes joint attention by aligning head gaze with the 3D target and executing a brief, safe arm‑pointing gesture (see \autoref{fig:overview}, red line). The robot maintains an appropriate standoff distance and, if necessary, repositions to ensure a clear line of sight.


\paragraph{Verbal Feedback} Throughout the task completion, the robot provides adaptive, detailed guidance with the steps the user has to follow (e.g., "Time to take your medicine, follow me!" / "Looking for your medicine bottle"). The robot then provides concise, sequential instructions (e.g., "Open the bottle" $\rightarrow$ "Take the prescribed number of pills" $\rightarrow$ "Drink water"). Each step elicits an explicit confirmation or offers a repeat/rephrase on timeout. 

Robot-user interaction is mediated by an LLM (GPT-4o) with a dialogue policy that balances task progress and safety to provide dementia-appropriate support during medication management. After navigating to the medication ROI, the robot verifies visual detection, delivers adaptive step-by-step instructions, and requests a final confirmation of intake before ending the routine (see \autoref{fig:overview}, purple line). The policy provides necessary guidance while minimizing distractions during execution. Prompts are engineered to handle edge cases (e.g., the user becomes distracted or expresses reluctance), to adapt tone and pacing to individual preferences, and to maintain a safe, supportive style that keeps users engaged and motivated to complete the task.

Across phases, audio and embodied cues are intentionally \emph{redundant yet complementary}: speech conveys intent and timing and embodied behaviors (gaze, pointing, proxemics) ground the dialogue in the physical scene. This event‑driven policy, tied to navigation and perception events, is designed to reduce working‑memory demands and cognitive workload, maintain joint attention, and support successful task completion while preserving user autonomy—an essential factor for effective robotic interaction and assistance for people with dementia.

\subsection{Gaze Sensing and Interaction Logging}
A lightweight, glasses-style eye tracker ETVision was worn by participants throughout the medication-taking task to unobtrusively capture gaze direction and fixation duration at 180 Hz. This allowed us to observe moments when participants looked at key objects (such as the pill bottle or robot interface) but did not act immediately, which may indicate confusion or difficulty \cite{yuan2022assessment}. Each participant completed a brief calibration procedure at the start of the session to ensure accurate gaze mapping. The resulting gaze data were time-stamped and synchronized with the robot’s verbal and gestural outputs to explore the feasibility of detecting potential ``confusion events.” While the present system does not yet act on gaze data in real time, data from this work will inform our planned gaze-triggered prompting feature.


\section{Formative Usability Study with Healthy Adults}

Before recruiting older adults and people living with dementia (PLWDs), we conducted a formative usability study with healthy adults to de‑risk the protocol and refine system components. This preliminary phase was motivated by practical and ethical considerations: to surface usability issues prior to testing with vulnerable participants, reduce the likelihood of confusion during task execution, and identify failure modes in robot‑assisted medication management that could adversely affect user experience or system performance. We treat this phase as formative rather than confirmatory; its purpose was feasibility and safety tuning, not generalization to the target population. We acknowledge the significant differences between this initial group and PLWDs, and we do not assume that findings fully extrapolate to the target population.

This study was approved by the Institutional Review Board (IRB \#\#\#\#). Inclusion criteria were: (1) age $\geq$ 18 years and (2) English - speaking. We recruited $N=12$ adults (ages 18–64) from the university. Six participants reported familiarity with dementia caregiving, and eight reported familiarity with similar technologies prior to participation. All sessions were conducted in a university laboratory setting. Insights from this phase informed subsequent refinements to the protocol and system components prior to recruiting older adults and PLWDs.

\subsection{Experimental Procedure and Task Design}

Prior to any study activities, participants reviewed and signed an informed consent form authorizing recording and the use of collected study data for academic research. Participants were reminded that participation was voluntary and that they could withdraw at any time if they felt discomfort. After consent, a pre‑study survey captured demographics (age, gender, education), background experience with technologies, and prior knowledge or experience with dementia caregiving. The experimenter then assisted each participant in wearing the ETVision gaze tracker and provided instructions describing the session flow.
\begin{figure}[!htp]
  \centering
  \includegraphics[width=0.6\columnwidth]{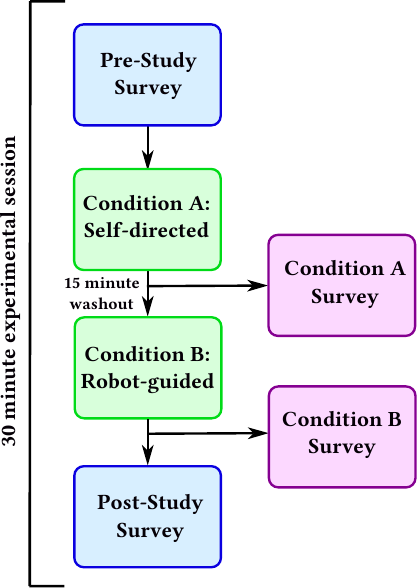}
\caption{Experimental procedure}
 \label{fig:exp}
\end{figure}
We employed a within‑session, two‑condition design for a simulated medication‑taking scenario (see \autoref{fig:exp}), separated by a 15‑minute washout interval, during which the environment was altered. Before each condition, we randomized the pill bottle’s location within a cluttered layout and reconfigured surrounding objects. This was intended to emulate the situational unfamiliarity that PLWDs often experience—even in otherwise familiar spaces—due to memory impairments. The two task conditions were: (A) a self‑directed baseline with optional, on‑request verbal guidance; and (B) a robot‑guided condition using our full navigation–perception–interaction pipeline. In both conditions, the goal was to locate a pill bottle prepared for the study (emptied) and open the cap; opening the cap marked task completion. All sessions were conducted in a university laboratory.

\paragraph{Condition A: Baseline (self‑directed).}
Participants navigated the lab and searched for the pill bottle independently. The robot remained present but passive, providing only brief verbal guidance on request (e.g., possible locations) and occasional encouragement. Guidance could be requested by pressing the \emph{Start Recording} button on the robot’s tablet to speak to the system. Upon completing the task, participants filled out a brief session questionnaire.

\begin{figure}[!htp]
  \centering
  \includegraphics[width=\columnwidth]{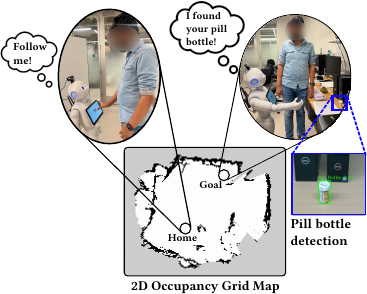}
\caption{The 2D occupancy grid map represents the lab used for the tests. Image on top left shows the interaction between the participant and the robot. Image on top right shows the interaction between the participant and the robot after it guided and detected the pill bottle.}
 \label{fig:map}
\end{figure}

\paragraph{Condition B: Robot‑guided. (\autoref{fig:map})}
Participants then repeated the activity with robot guidance. The tablet displayed two controls: \emph{Start Navigation} and \emph{Start Recording}. As demonstrated in \autoref{fig:map}, when \emph{Start Navigation} was pressed, the robot escorted the participant toward a predefined region of interest (ROI) in the lab. After the bottle was detected, the robot oriented its base and head and executed a short pointing gesture to establish joint attention, then prompted the participant to press \emph{Start Recording} to begin the interaction phase. Once the participant indicated that the medication task was complete (i.e., the cap was opened), the robot acknowledged completion and terminated the routine.

\subsection{Evaluation Metrics}
We assessed the system through three complementary approaches to evaluate workload, efficiency, and overall user experience.

\paragraph{Workload Assessment (NASA-TLX)} 
Participants completed the NASA Task Load Index (NASA-TLX) after each condition to assess perceived workload across six dimensions: Mental Demand, Physical Demand, Temporal Demand, Performance, Effort, and Frustration. Items were rated on a 1–10 scale. As pairwise weights were not collected, we computed Raw TLX as the unweighted mean of the six dimensions and linearly transformed scores to a 0–100 scale for interpretability: 
\[
\mathrm{TLX}_{0\text{--}100}=\frac{\mathrm{TLX}_{\mathrm{raw}}-1}{10-1}\times 100.
\]
This approach enabled within-participant comparison across assistance levels to determine how cognitive load varied with our adaptive framework.

\paragraph{Behavioral Efficiency (Gaze-Tracking)} 
Eye-tracking data provided objective measures of task performance: (1) time-to-locate-duration from room entry to first verified fixation on the correct pill bottle-and (2) robot-interaction rounds-number of conversational exchanges between participant and robot. These metrics captured both task efficiency and the extent of human-robot engagement.

\paragraph{Perceived Usability (Adapted SUS/QUIS)} 
A post-study questionnaire comprising nine items adapted from the System Usability Scale (SUS)~\cite{norman1998questionnaire} and Questionnaire for User Interaction Satisfaction (QUIS)~\cite{brooke1996sus} assessed perceived ease of use, effectiveness, and satisfaction. As this was a formative pilot, we used an adapted subset rather than complete instruments. Items were rated on a five-point Likert scale (1 = strongly disagree, 5 = strongly agree), with items Q2 and Q4 reverse-coded. We computed an adapted usability composite by averaging item scores at the participant level and transforming to a 0–100 scale for consistency with other metrics. We report means with 95\% confidence intervals, medians, interquartile ranges, and Cronbach's alpha for internal consistency.

\paragraph{Qualitative Feedback} 
Open-ended questions elicited concerns and suggestions (e.g., ``One thing about the robot that needs improvement"; ``Any concerns about privacy or safety while using the robot for medication management") to inform iterative design and refinement.

\subsection{Ethics, Privacy, and Data Management}
All participants provided written informed consent and could withdraw at any time without penalty. Gaze‑tracking data were de-identified at the point of collection and stored without direct identifiers. Sessions were video‑recorded on a GoPro and held temporarily on the device; audio recordings were deleted immediately after each interaction. After each session, video files were transferred to a secure, access‑restricted hard drive available only to IRB‑approved study personnel. All identifiable information was handled confidentially in accordance with the consent form and institutional policy.

\section{Results and Discussion}
\label{sec:results}

\begin{figure}[!htp]
  \centering
  \includegraphics[width=0.6\columnwidth]{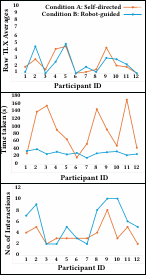}
  \caption{Top plot indicates the mean Raw TLX scores for each participant for each condition. Middle plot shows the time taken by each participant to find the pill bottle for each condition. Bottom plot shows the number of interactions each participant had with the robot for each condition. }
  \label{fig:tlx}
\end{figure}

Our formative evaluation with healthy adults ($N=12$ , 9 aged 20-30 and 3 aged 30-60) examined whether the adaptive multimodal framework successfully balanced assistance provision with cognitive demand, and how users perceived the system's usability. Half of the participants (6/12) reported familiarity with people living with dementia or direct dementia caregiving experience. \autoref{fig:tlx} displays individual participant data for workload, task completion time, and robot interaction frequency across both conditions (Condition A: verbal only; Condition B: verbal+gesture+navigation). Results are organized around three key findings that inform future deployment with PLWDs.

\subsection{Adaptive Robotic Assistance Reduces Workload Without Overwhelming Users}
Raw NASA–TLX scores were uniformly low across conditions (\autoref{tab:nasatlx}), with mean normalized values of 20.24/100 for Condition A (verbal only) and 13.89/100 for Condition B (verbal + gesture + navigation). On this 0–100 scale, lower scores indicate lower perceived workload, suggesting a lighter cognitive burden in the embodied condition. While the richer multimodal condition showed numerically lower workload, the small difference and restricted variance (both <25/100) suggest floor effects in this controlled laboratory setting with healthy participants performing relatively simple tasks.
\begin{table}[!htp]
  \centering
  \caption{Task-level summary: Raw NASA--TLX (0--100), time-to-locate (s), and robot-interaction rounds.}
  \label{tab:nasatlx}
  \begin{adjustbox}{width=\columnwidth}
    \begin{tabular}{lccc}
      \toprule
      Condition & TLX (M) (0--100) & Time-to-locate (s) & Robot-interaction rounds \\
      \midrule
      Condition A (verbal only)        & 20.24 & 91.8 & 3 \\
      Condition B (verbal+gesture+nav) & \textbf{13.89} & \textbf{29.7} & \textbf{5.5} \\
      \bottomrule
    \end{tabular}
  \end{adjustbox}
\end{table}

The low workload across conditions is encouraging: even the most comprehensive assistance level (Condition B) did not overwhelm users cognitively. This suggests that multimodal guidance-combining verbal instructions, gestural cues, and physical navigation-can be delivered without imposing excessive cognitive burden. However, the floor effect limits our ability to detect meaningful differences. Future studies with PLWDs performing more challenging medication regimens in naturalistic home environments will provide a more sensitive test of workload differences across assistance levels.

\subsection{Multimodal Robotic Guidance Enhances Task Efficiency and Engagement}
\begin{figure}[!htp]
  \centering
  \includegraphics[width=\columnwidth]{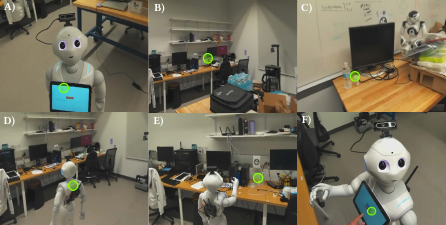}
  \caption{Top row indicates the data collected during Condition A. Bottom row indicates the data collected during Condition B. A) Participant interacts with the robot. B) Participant looks around the room to search for the pill bottle. C) Participant finds the pill bottle. D) Participant is following the robot to pill bottle location. E) Robot gestures towards the pill bottle. F) Participant interacts with robot for further guidance.}
  \label{fig:gaze}
\end{figure}

Gaze-tracking data revealed substantial practical benefits of embodied assistance (\autoref{fig:gaze}). Participants located target medications three times faster with full multimodal support (29.7s) compared to verbal-only reminders (91.8s). Correspondingly, robot-interaction rounds increased from 3 (Condition A) to 5.5 (Condition B), indicating that participants engaged more extensively with the robot when richer guidance was available.

These behavioral measures demonstrate that adaptive escalation from minimal to comprehensive assistance produces tangible efficiency gains. For PLWDs who experience spatial disorientation or difficulty executing multi-step tasks \cite{Puthusseryppady2020}, faster medication location and more supported interactions, along with step-by-step guidance, could meaningfully impact adherence and reduce caregiver burden. The increased engagement in the multimodal condition aligns with our design goal of providing "just-enough" assistance: users naturally sought more interaction when they received navigational and gestural support, suggesting the system successfully scaffolded task completion rather than replacing user agency.

\subsection{High Perceived Usability with Interface Refinement Needs}
\begin{figure}[!htp]
  \centering
  \includegraphics[width=\columnwidth]{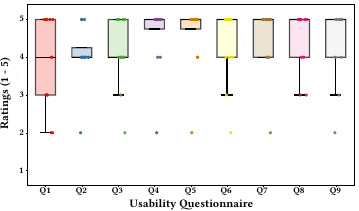}
  \caption{Adapted usability questionnaire: item distributions and composite (0--100).}
  \label{fig:use}
\end{figure}

The adapted usability composite yielded M=83.10 (95\% CI [70.39, 95.82]), median=88.89, IQR=6.25, with high internal consistency ($\alpha=0.954$). While these scores cannot be directly compared to standard SUS/QUIS benchmarks due to our adapted instrument, the high ratings and tight clustering suggest strong perceived usability. Figure~\ref{fig:use} shows item-level distributions, revealing consistently positive responses across all nine questions, with most items showing median scores at or above 4 ("somewhat agree") on the 5-point scale. The reversed items (Q2 and Q4) showed appropriately low scores, indicating participants did not perceive the system as complex or cumbersome.

Qualitative feedback was broadly positive, with participants and stakeholders noting the robot's potential to support PLWDs during their daily routine activities, promoting their quality of life and independence. However, one consistent concern emerged: confusion around the tablet's Start/Stop Recording buttons used to initiate interaction. This interface element requires redesign to ensure intuitive operation, particularly for users with cognitive impairments.

The strong usability ratings provide preliminary validation that our adaptive framework is comprehensible and acceptable to users. The high internal consistency ($\alpha\approx0.95$) indicates our adapted questionnaire reliably captured a coherent usability construct. However, the identified confusion with interaction initiation highlights a critical design consideration for PLWD deployment: interface elements must be unambiguous and cognitively minimal. We plan to replace the current button system with more intuitive triggers (e.g., proximity-based activation or voice-initiated commands) in the next iteration.

\subsection{Limitations and Future Work}
This pilot ($N=12$) with healthy adults in a single-room laboratory offers limited statistical power and generalizability to older adults and people with mild cognitive impairment; the simplified setting likely introduced floor effects. Participants reported slow response times and delays in object detection/interaction, but because the perception--planning--actuation pipeline was not instrumented end-to-end, latency could not be attributed to specific components (e.g., perception, path planning, UI feedback). Finally, workload and usability were assessed using Raw NASA--TLX (unweighted) and an adapted SUS/QUIS subset, so composites should be interpreted as internal measures rather than canonical benchmarks.

We will scale to an adequately powered study that includes older adults and people with mild cognitive impairment and extends evaluation beyond a single-room lab to multi-room home/clinic layouts with occlusions and dynamic obstacles. System priorities include end-to-end latency instrumentation; strengthened motion planning (dynamic obstacle avoidance, narrow-corridor navigation, human-aware pathing, and semantic/topological maps with rapid re-planning under perception dropouts) \cite{chatrola2025multi}; privacy-preserving visual monitoring for task state (on-device inference, event-level logs in place of raw video, configurable retention, and explicit consent/UI indicators); and an expanded tablet UI (stepwise cards with large icons, progress/status and error-transparency, multimodal redundancy, and accessible design). We will report effect sizes with confidence intervals and conduct an a priori power analysis to guide sampling.

\vspace{-3mm}
\section{Conclusion}
We presented an adaptive, multimodal robotic framework operationalizing occupational therapy’s \emph{assist-as-needed} principle for medication management on a social robot platform. The system links perception to action and dialogue via a unified pipeline: 
YOLO v11-based detection with depth-guided refinement to recover a 3-D target; joint attention via head alignment and a brief pointing gesture; push-to-talk speech transcription (Whisper); LLM-driven stepwise guidance (GPT‑4o); and an event-driven policy escalating from reminders to full multimodal support only as needed—preserving autonomy while maintaining task progress. In a formative lab study with healthy adults and dementia-care stakeholders, participants reported high perceived usability on our adapted instrument (M $\approx83/100$; $\alpha\approx0.95$); gaze-derived timing suggested faster target acquisition and greater engagement under richer embodied guidance, while workload remained low across conditions. Open-ended feedback highlighted UI confusion around \texttt{Start/Stop Recording} and perceived latency.
These results, while preliminary, indicate feasibility and promise for graded, embodied assistance in medication routines. Limitations include the small sample, controlled setting, and use of adapted measures; accordingly, we plan powered studies with older adults and PLWDs in naturalistic home environments, end-to-end latency instrumentation, strengthened navigation/planning, privacy-preserving visual monitoring for task state, and an expanded, accessible tablet UI.
Taken together, this work establishes a theoretically grounded, empirically informed foundation for person-centered, autonomy-preserving robotic assistance in dementia care. By translating occupational therapy principles into adaptive HRI design, our framework advances toward robotic systems that complement rather than replace human caregiving—ultimately empowering independence and dignity for people living with dementia.

\bibliographystyle{ACM-Reference-Format}
\bibliography{references}

\newpage
\FloatBarrier
\appendix
\section{Survey Instruments}
\subsection{NASA--TLX (condition-based workload)}
\label{app:nasa_tlx}
This appendix reproduces the exact NASA--TLX items used after each task to enable within-participant comparison across assistance types. Items were rated on a 1--10 scale (1 = low, 10 = high). Because pairwise weights were not collected, we computed \emph{Raw TLX} as the unweighted mean of the six items \cite{hart1988development} and mapped scores to a 0--100 display scale:
\begin{equation}
\mathrm{TLX}_{0\text{--}100}=\frac{\mathrm{TLX}_{\mathrm{raw}}-1}{10-1}\times 100.
\label{eq:tlx_map}
\end{equation}

\begin{table}[H]
  \centering
    \caption{NASA--TLX item wordings and dimensions. Administered after each condition.}
    \label{tab:nasa_tlx_items}
    \setlength{\tabcolsep}{4pt}
    \renewcommand{\arraystretch}{1.15}
    \begin{tabularx}{\columnwidth}{p{5cm} p{3cm}}
      \toprule
      \textbf{Item wording} & \textbf{Dimension} \\
      \midrule
      How mentally demanding was the task? & Mental Load \\
      How physically demanding was the task? & Physical Load \\
      How hurried or rushed was the pace of the task? & Temporal Load \\
      How successful were you in accomplishing what you were asked to do? & Performance \\
      How hard did you have to work to accomplish your level of performance? & Effort \\
      How insecure, discouraged, irritated, stressed, and annoyed did you feel? & Frustration \\
      \bottomrule
    \end{tabularx}
\end{table}

\subsection{Adapted usability questionnaire (post‑study)}
\label{app:usability}
This appendix lists the nine post‑study items adapted from SUS and QUIS. Items were answered on a five‑point Likert scale (\textit{strongly disagree} to \textit{strongly agree}). Following standard keying, Q2 and Q4 were reverse‑coded prior to aggregation. We report an \emph{adapted usability composite} formed as the participant‑level mean of all items, mapped to 0--100 for display:
\begin{equation}
U_{0\text{--}100}=\frac{U_{\mathrm{raw}}-1}{5-1}\times 100,
\label{eq:usability_map}
\end{equation}
where $U_{\mathrm{raw}}$ is the 1--5 mean. Because this is an adapted subset, values are presented as an internal composite and are not compared to canonical SUS/QUIS norms \cite{brooke1996sus,norman1998questionnaire}.

\begin{table}[H]
  \centering
    \caption{Adapted SUS/QUIS items and mapped constructs (administered once post‑study).}
    \label{tab:usability_items}
    \setlength{\tabcolsep}{4pt}
    \renewcommand{\arraystretch}{1.15}
    \begin{tabularx}{\columnwidth}{p{0.5cm} p{5cm} p{2.3cm}}
      \toprule
      \textbf{ID} & \textbf{Item wording} & \textbf{Construct} \\
      \midrule
      Q1 & I think that I would like to use the robot frequently. & Usability \\
      Q2 & I found the assistance with the robot unnecessarily complex. & Usability (reverse‑coded) \\
      Q3 & I thought the robot was easy to use. & Perceived ease of use \\
      Q4 & I think that I would need the support of a technical person to be able to use the robot. & Perceived ease of use (reverse‑coded) \\
      Q5 & I would imagine that most people would learn to use the robot very quickly. & Perceived ease of use \\
      Q6 & I felt very confident using the robot. & Anxiety \\
      Q7 & It is easy to understand the robot's instructions and know what to do. & Information Quality \\
      Q8 & I would follow the advice the robot gives me. & Trust \\
      Q9 & Overall, I am satisfied with the robot. & Overall Satisfaction \\
      \bottomrule
    \end{tabularx}
    \begin{tablenotes}[flushleft]
      \footnotesize
      \item Note: Reliability and descriptive statistics for these adapted items are reported in the main text; item‑level distributions are shown in Fig.~\ref{fig:use}.
    \end{tablenotes}

\end{table}


\end{document}